\RequirePackage{fix-cm}
\documentclass[twocolumn]{svjour3}          
\smartqed  
\usepackage{graphicx}
\usepackage{amsmath}
\usepackage{amssymb}
\usepackage{booktabs}
\usepackage{subfigure}

\usepackage{multirow}

\usepackage{color}

\usepackage[pagebackref,breaklinks,colorlinks]{hyperref}

\usepackage{natbib}
\setcitestyle{authoryear,open={(},close={)}}

\begin{document}

\title{When A Conventional Filter Meets Deep Learning: Basis Composition Learning on Image Filters}

\titlerunning{When A Conventional Filter Meets Deep Learning: Basis Composition Learning on Image Filters}

\author{Fu Lee Wang  \and Yidan Feng
         \and Haoran Xie
         \and Gary Cheng
         \and Mingqiang Wei
}

\institute{Fu Lee Wang \at Hong Kong Metropolitan University \\
Yidan Feng and Mingqiang Wei \at  Nanjing University of Aeronautics and Astronautics \\
Haoran Xie \at Lingnan University\\
Gary Cheng  \at Education University of Hong Kong
%
}
\date{Received: date / Accepted: date}

\maketitle

\begin{abstract}
\sloppy{  
Image filters are fast, lightweight and effective, which make these conventional wisdoms preferable as basic tools in vision tasks. In practical scenarios, users have to tweak parameters multiple times to obtain satisfied results. This inconvenience heavily discounts the efficiency and user experience. We propose basis composition learning on single image filters to automatically determine their optimal formulas. The feasibility is based on a two-step strategy: first, we build a set of filtered basis (FB) consisting of approximations under selected parameter configurations; second, a dual-branch composition module is proposed to learn how the candidates in FB are combined to better approximate the target image. Our method is simple yet effective in practice; it renders filters to be user-friendly and benefits fundamental low-level vision problems including denoising, deraining and texture removal. Extensive experiments demonstrate that our method achieves an appropriate balance among the performance, time complexity and memory efficiency. The code is available at: \href{https://github.com/Dengsgithub/composition-learning}{https://github.com/Dengsgithub/composition-learning}.
}
\end{abstract}

\section{Introduction} \label{sec1}
Filtering is a fundamental operation in image processing. Image filters benefit immense applications such as image denoising \cite{fil_denoise,fil_denoise2}, texture removal \cite{Cho2013_BTF,RGF}, edge extraction \cite{fil_edge1,fil_edge2,fil_edge3} and deraining \cite{Xu2012a_derain,Zheng2013_derain}, to name a few. Parts of which are now dominated by CNN-based methods \cite{DLdenoise1,DLdenoise2,DLderain1,DLSmooth,DLderain2,ijcai/DengFW0C0Z021,corr/abs-2112-06451}. Although achieving better performance, there are tons of parameters to optimize for image-to-image translation tasks in CNN-based methods and the training process consumes massive data to avoid overfitting, which is an overkill to basic pre-processing tasks. In contrast, filters are not only fast and lightweight, but also flexible and extendable. There are mature algorithms \cite{BF_ac1,ac2,guidedfil_ac} to accelerate classical filters and recent studies embed filters with domain-specific knowledge to render them more powerful \cite{Cho2013_BTF,Jeon2016}. There is no doubt that conventional wisdoms remain vital as a basic tool in practical applications, especially in resource-limited circumstances.

\begin{figure}[t]
	\centering
	\subfigure[Histogram of tuning trials]{
		\includegraphics[width=0.48\linewidth, height=0.36\linewidth]{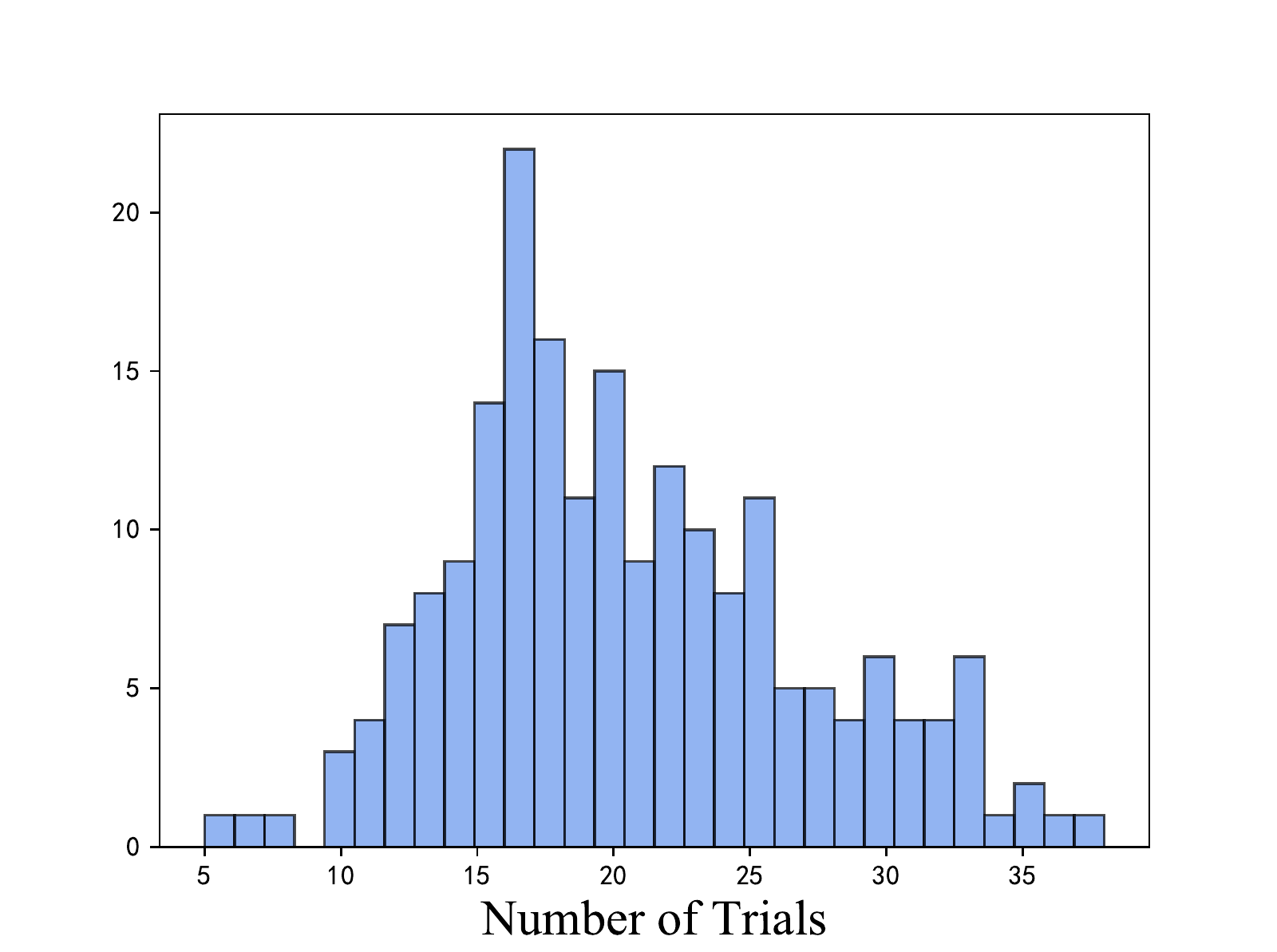}}
	\subfigure[Best filtering results]{
		\includegraphics[width=0.48\linewidth, height=0.36\linewidth]{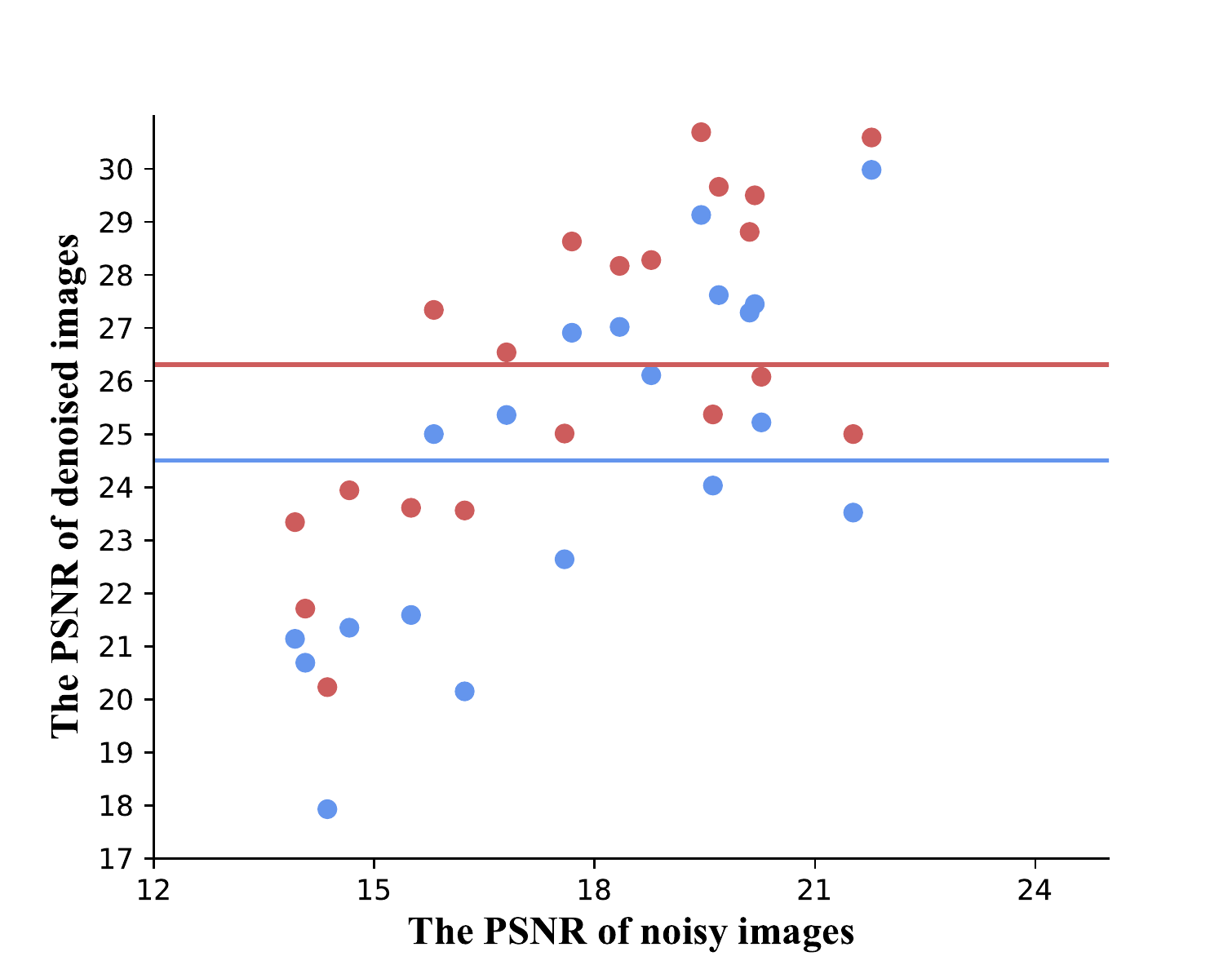}}
	\caption{Results of user study to figure out the real cost and effect in bilateral filtering. 20 participants are asked to filter 10 images of noise levels from 25 to 50. The number of trials and the best filtering results are recorded  (blue: user average; red: ours). }
	\label{userHist}
\end{figure}

However, there lies a shared and inherent problem in applying filters, that is, the tedious parameter adjustment. Obtaining desired results relies on continually tweaking parameters to accommodate various inputs. This process is arduous and burdensome, since it is hard to know whether the current result is optimal before several more trials. Moreover, one has to coordinate among several parameters when applying complex filters, which could be confusing for un-experienced users. 
In real applications, it is impossible to tune the parameters for thousands of times, nor fix the thing just in one trial. To find out the real cost and effect in applying the bilateral filter, we conduct a user study with 20 participants asked to filter 10 images of noise levels from 25 to 50. The number of trials and the best filtering results are recorded in Fig. \ref{userHist}. Obviously, it actually costs much more to filter out a satisfactory result and that result might not represent the best potential of the filter.

In this paper, we look at this intriguing problem: can users apply filters without the laborious parameter tuning process and automatically achieve the optimal filtered result? We note that there are efforts in designing adaptive filters \cite{Foi2007,Jeon2016,fil_denoise,median_adaptive,cvpr/YinGQ19,aaai/LiuZLHY020} to cope with various inputs, however, they generally introduce additional parameters to set thresholds or to measure local conditions, instead of reducing the number of parameters. Meanwhile, this adaptability relies on the specific characteristics of some given filters, which makes them difficult to be generalized to other cases.
These problems motivate us to propose basis composition learning for single image filters, which provides a lightweight yet effective solution to many image-to-image translation tasks. We solve the problem of optimal filtering within two steps: first, build a candidate pool which contains qualified approximations robust to varying inputs; second, search for the best combination in this limited and quantified space. Namely, the optimal result is $G = \phi\left(G_{1},G_{2},\cdots,G_{n}\right)$, where $\phi$ can be a linear or nonlinear function. In this way, we translate the problem into predicting the optimal composition of a certain group of filtered images, which we call them `filtered basis' (FB). Specifically, we construct FB by filtering the input under several groups of parameters. Note that this process also requires some tuning experience but these efforts are paid in an offline stage, which makes it user-friendly in the runtime stage. After forming a proper FB to express the target space, we adopt very shallow fully connected layers to learn the function $\phi$.

Fully connected (FC) networks can theoretically imitate any nonlinear transformation, but it fails to maintain the spatial structure in image translation tasks. Instead, convolutional layers are applied as trade-offs between performance and computational cost, but they learn only local spatial features and rely on the FC structure to connect the embeddings. In our method, FC layers are directly applied along the channels to blend image-level features produced by parameterized image filters. In addition, we propose to enhance this framework by an extra branch learning from FB's residuals, which improves the performance by feature expansion with negligible cost.

\textit{Our goal is to provide a general basis composition learning paradigm for image filters, rather than being the top in specific applications.} In summary, our contributions are:\\
1) We for the first time propose the concept of basis composition learning for image filters, which provides a novel perspective to combine domain knowledge with learning schemes. The proposed framework renders filters to be user-friendly, since the burden of parameter tuning is transferred to the construction of filtered basis and composition learning in an offline stage.\\
2) To fully exploit residual features, we present a dual-branch composition network based on fully connected layers, which is effective in processing multi-channel inputs.\\
3) Popular filters such as median filter, bilateral filter and rolling guidance filter are implemented under the proposed framework. We show that composited filters are competitive compared with previous methods in a variety of applications including denoising, deraining and texture removal.

The rest of this paper is organized as follows: In Section \ref{sec2}, we give a brief introduction of literature pertaining to image filters and FC networks for images. After that, we present our model and explain how to construct FB in Section \ref{sec3}. The experimental results are shown in Section \ref{sec4} and more issues are discussed in Section \ref{sec5}. Finally, we make conclusion in Section \ref{sec6}.

\begin{figure*}[]
	\centering
	\includegraphics[width=1.0\textwidth]{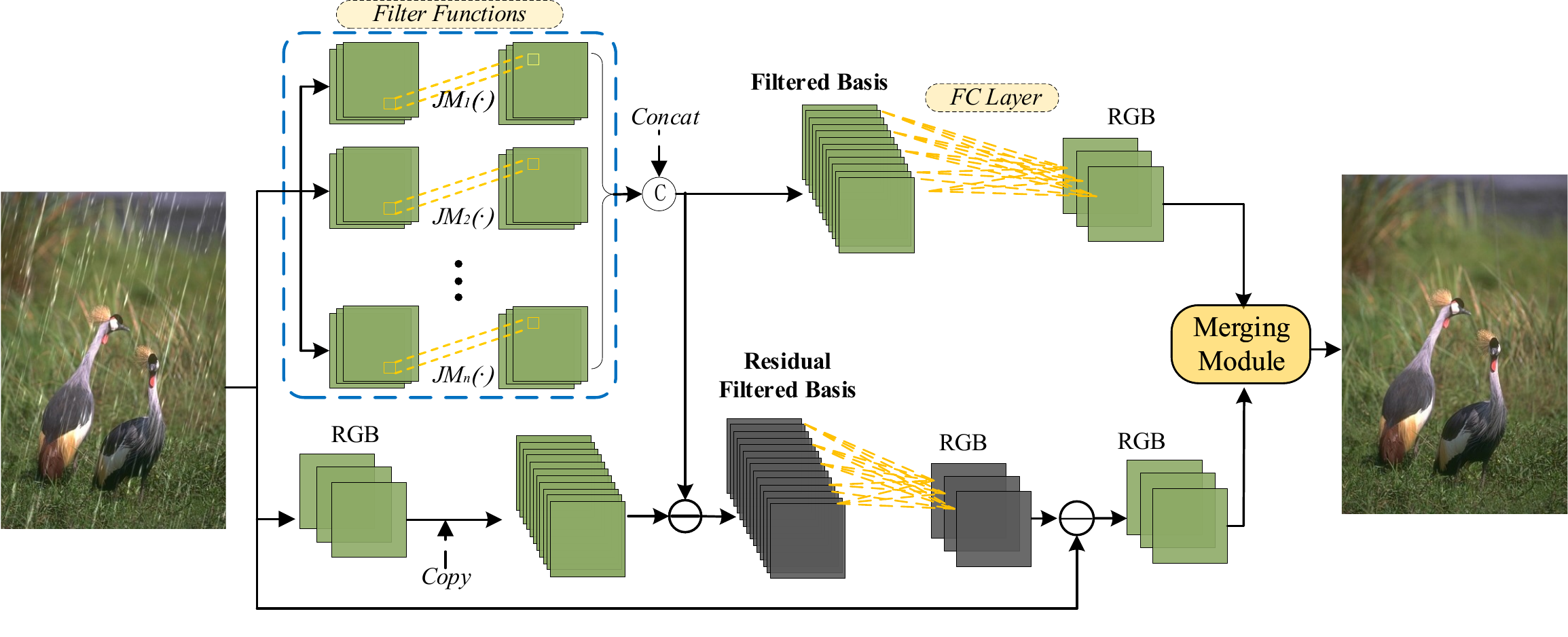}
	\caption{What will happen when a conventional filter meets deep learning? The proposed image filtering framework that combines domain knowledge and deep learning can tell you the answer. The proposed basis composition learning paradigm for conventional image filters provides a lightweight yet effective solution to many image-to-image translation tasks. The problem of optimal filtering is solved by two steps: first, build a candidate pool which contains qualified approximations robust to varying inputs; second, search for the best combination in this limited and quantized space. In this way, the problem of optimal filtering is translated into predicting the optimal composition of a certain group of filtered images. In addition, to fully exploit residual features, we present a dual-branch composition network based on very shallow yet fully connected layers, which is effective in processing multi-channel inputs.}
	\label{pipeline}
\end{figure*}

\section{Related Work} \label{sec2}

\textbf{Image Filters.}
%
The studies on image filters are mature and long-standing. Various filters have been developed for many purposes. According to the operation properties, image filters can be divided into linear filters and non-linear filters. Representative linear filters include box filter, Gaussian filter and Sobel filter, etc. Linear shift invariant filters (LSI) are theoretically robust and have ideal properties for signal analysis, but fail to decompose the correlated signal and noise. In later studies, non-linear filters \cite{nonlinear1,nonlinear2,nonlinear3} are proposed to take advantage of specific statistical properties of the signal, which boost the performance of filters in various applications.

Bilateral filter \cite{Tomasi1998} benefits edge-aware smoothing and image denoising by weighting both spatial and photometric distances, and is able to cope with more complex tasks combined with a guidance image, such as texture removal \cite{RGF,Cho2013_BTF} and dehazing \cite{JBF_dehaze}. Guided filter \cite{GuidedFil} provides another framework to embed well-engineered guidance images, therefore incentivizes plenty applications, including multi-scale demosaiching \cite{guidedfil_demosaiching}, deraining \cite{Zheng2013_derain} and image fusion \cite{guidedfil_imagefusion}. Compared with nonlinear mean filters, median filter and its modifications \cite{median_adaptive,ac3} perform well in removing impulse noise. By calculating gradients and setting threshold, Sobel filter \cite{sobelfil} and Canny filter \cite{canny_fast} produce a binary image to classify edges. There are also accelerated versions of these well-studied filters \cite{BF_ac1,guidedfil_ac,ac4,ac3,ac2} to ensure their efficiency.

To improve the performance of filters, adaptive filters \cite{Foi2007,Jeon2016,median_adaptive} are designed to suit specific tasks. In particular, a later work \cite{Jeon2016} aims to remove textures while preserving the structural edges and corners by controlling the smoothing strength for guidance image generation. To filter with adaptive kernel scales, they additionally introduce a $\sigma_{e}$ to control the flatness transition and a threshold $\sigma$ to define the maximum scale of texture to be removed.

Recent advances in this field allow to decompose an image into fine-grained layers \cite{filrec1,filrec2,filrec3,Cho2013_BTF}. Compared with optimization-based and CNN-based methods, filters are flexible, fast and lightweight, but commonly suffer from tedious parameter tuning.

\textbf{Fully Connected Networks for Images.}
Due to the prohibitive computation complexity, it is intractable to directly apply fully connected (FC) networks by reshaping an image to a feature vector. To alleviate it, extant approaches attempt to shrink the input size either by forming an image descriptor capturing image characteristics or by splitting the image into patches. Driss et al. \cite{MLP_descriptor} train an MLP to recognize characters which are represented by a unified character descriptor. Foody \cite{MLP_descriptor2} feeds the MLP with inputs associated with spectral wavebands to test the classification accuracy. Similarly, Zhang et al. \cite{CNN_MLP1} propose to substitute CNN with MLP when the CNN results are of low confidence, where the MLP is supplied with GLCM texture features including the Mean, Variance, Homogeneity, etc. However, these descriptors are hard to be applied in image-to-image translation tasks, since descriptors lose context and appearance information required to reconstruct an image.

In contrast, patch-based MLP methods \cite{MLP_patch1,MLP_patch2,MLP_patch3} for image denoising learn a mapping from patch to patch, which completely retains the local information from the original image. Consequently, errors are introduced when aggregating patches to reconstruct the final image. Actually, small patches sacrifice non-local characteristics to improve efficiency. Recent advances focus on combining the merits of both CNNs and FC networks. For instance, Zhang et al. \cite{CNN_MLP2} find that the performance of the MLPs and the CNNs partly relies on the signal distribution of the image patches. In particular, MLPs retain much more details and the CNNs give a much smoother result. Thus, the final results are obtained as a weighted average of the two patches estimated from the CNN branch and the MLP branch. In summary, when applying FC networks to tackle images, extant methods sacrifice accuracy for limited resources.

\section{Basis Composition Learning} \label{sec3}
Considering a filter $J_{M_{i}}$ with tunable parameters $M_{i}$, the filtering results vary with different settings of $M_{i}$ to a certain extent. This variation in essence provides a range of approximations covering various input situations. To express the space of optimal solutions, we sample this continuous variation to form the set of `filtered basis' (FB), written as $\mathbf{F}=\left\{{J}_{M_{i}}(O)|i=1,2,...,n\right\}$.
Our goal is to learn the best blending of the filtered basis to produce the optimal solution $G_{c}$. If denoting the blending function as $\phi(\cdot)$, the process can be simply expressed as
\begin{equation}
	G_{c}=\phi\left(\mathbf{F}\right)
\end{equation}
and the loss function can be formulated as
\begin{equation}
	\mathcal{L}_{c}=\mathcal{L}\left(\phi\left(\mathbf{F}\right), \mathbf{G}_{c}\right)
\end{equation}
where $\phi(\cdot)$ is defined as a shallow fully connected network to weigh different channels. In most applications, we empirically demonstrate that a single-layer FC, which equals to linear regression, is enough to composite the FB.

Under this basic scheme, we further propose a dual-branch composition architecture. As shown in Fig. \ref{pipeline}, the composition module consists of a content learning branch and a residual learning branch, which simultaneously learns the target image and the ground-truth residual. The objective function of the residual learning branch is written as
\begin{equation}
	\mathcal{L}_{r}=\mathcal{L}\left(\varphi\left(\mathbf{R}\right), \mathbf{G}_{n}\right)
\end{equation}
where $\mathbf{R}=\left\{O-{J}_{i}(O)|i=1,2,...,n\right\}$ is the set of residual features and $\mathbf{G}_{n}$ denotes the ground-truth artifact (e.g., noise, textures, and rain streaks). $\varphi(\cdot)$ represents the residual learning network which shares the same structure with the content learning branch. Note that the purpose of applying this dual-branch representation is to take full advantage of our filtered basis as features. Once the content features are obtained from conventional filters, we are able to leverage the corresponding residual features to constrain our model from the opposite perspective. Taking image denoising as an example, it is intuitive that the noise-to-noise residual learning branch emphasizes more on noise-invariant features, while the clean-to-clean content learning branch benefits more on preserving the image's original information. In other words, the estimated content $\phi(\mathbf{F})$ tends to retain more multi-scale features (e.g., structures, details) while another estimation $O-\varphi(\mathbf{R})$ tends to contain less noise. To construct the optimal solution, these two learned approximations are further composited by weighted channel blending and the loss function is written as
\begin{equation}
	\mathcal{L}_{m}=\mathcal{L}\left(\eta\left(\mathbf{M}\right), \mathbf{G}_{c}\right)
\end{equation}
where $\mathbf{M}=\left\{\phi(\mathbf{F}),O-\varphi(\mathbf{R})\right\}$ is the set composed of the output from the content learning branch and the residual of the output from another branch. $\eta(\cdot)$ refers to the weighted channel blending. Consequently the overall loss function is
\begin{equation} \label{loss_t}
	\mathcal{L}=\alpha \mathcal{L}_{c}+\lambda \mathcal{L}_{r}+\gamma \mathcal{L}_{m}
\end{equation}
where $\alpha$, $\lambda$ and $\gamma$ are trade-off weights set as 0.1, 0.1 and 1, respectively. We use a batch size of 1 and train the network for 250 epochs on a 1080Ti GPU. Moreover, we implement the models and evaluate all experiments in Pytorch, and adopt the Adam optimizer to train the model. The learning rate is initialized to 0.1 and is divided by 5 every 50 epochs.

\subsection{Analysis on Filtered Basis}
In our framework, Filtered Basis (FB) is defined as a set of approximations obtained from domain-specific filters under different parameter configurations. Compared with the models learned from only the original images, which first require sophisticated convolutional networks to learn effective features and then resort to fully connected structure for blending, the composition learning module accepts FB as input features and directly learns their weights along channels to construct the optimal solution. In this way, the problem has been translated to how to properly weigh the existing candidates rather than learning from scratch, which greatly eases the learning burden. As shown in Table \ref{tab:tab1}, given a simple network struture (in this case a single layer fully connected network), a one-dimensional FB boosts the performance of denoising and the performance grows when the magnitude of FB increases.

The choice of FB plays an important role in our method. FB should be not only highly related to the target but also robust to interference. These two properties ensure that the set of FB can accommodate qualified approximations from varying inputs. For the convenience of discussion, we take the popular bilateral filter for image denoising as an example to introduce the construction of FB, and other filters follow the similar scheme.

\textbf{Bilateral Filter.}
First we give a brief introduction of bilateral filter (BF). BF is a nonlinear edge-preserving denoiser which suppresses signals with similar intensity values or close positions. Given an input noisy image $I$, BF computes an output image $J$ by
\begin{equation}
	J(p)=\frac{1}{K_{p}} \sum_{q \in N(p)}f(\|q-p\|)g(\|I(q)-I(p)\|) I(q)
\end{equation}
where $N(p)$ is the set of neighboring pixels of the operating pixel $p$ and $K_{p}=\sum_{q \in N(p)} f(\|q-p\|)g(\|I(q)-I(p)\|)$ is the normalization term. The kernel functions $f(\cdot)$ and $g(\cdot)$ are typical Gaussian functions. Their standard deviations  $\sigma_{s}$ and $\sigma_{r}$ control the smoothing strength and the ability to preserve edges in a filtering window of size $k*k$. A filtering window with a larger $k$ provides more neighboring information that benefits the noise removal of different levels.

\textbf{Construction of FB.}
The factors determining the parameter configurations of FB are as follows:
\begin{enumerate}
	\item the effective range of each parameter
	\item the amount of adjustable parameters
	\item the parameter sampling scheme
\end{enumerate}
The first two factors resort to the properties of the specific filter and the empirical experience of parameter tuning. Specifically, the effective range of the three parameters in BF can be commonly decided as $\sigma_{s} \in (0.1, 1.1)$, $\sigma_{r} \in (0.5, 3.5)$ and $k = 15$. Then, it is intuitive to uniformly sample parameters in linear space and use their combinations as the final configurations. We call this process Direct Isometric Sampling (DIS). DIS is capable to cover both effectiveness and diversity, but different combinations generated in this way can yield very similar effects, which causes redundancy in FB. To alleviate this problem, we propose the Indirect Isometric Sampling (IIS). First, we apply DIS with a small interval to generate a number of candidates. Then, we isometrically sample the quantitative result of these candidates. Since the relative distance tends to be source-invariant, the corresponding parameter sets can be employed to build robust FB under different inputs. For instance, $\sigma_{s}$ and $\sigma_{r}$ are sampled with the interval of $0.1$ and $0.5$, generating $77$ candidates of $PSNR \in (20,45)$. If we take the interval of $5$ to sample the PSNR, six groups of parameters can be directly obtained for constructing the FBs.

\begin{table}[h]
	\footnotesize
	\centering{
		\caption{\small Quantitative evaluations under different settings of FB}
		\label{tab:tab1}
		\vspace{5pt}
		\begin{tabular}{|c|c|c|c|c|c|c|}
			\hline
			\multicolumn{2}{|c|}{Magnitude of FB} & 0                      & 1                      & 3     & 9     & 16    \\ \hline
			\multirow{2}{*}{$\sigma=25$}     & DIS    & \multirow{2}{*}{22.42} & \multirow{2}{*}{28.29} & 29.41 & 30.14 & 30.23 \\ \cline{2-2} \cline{5-7}
			& IIS    &                        &                        & 29.98 & 30.22 & 30.20 \\ \hline
			\multirow{2}{*}{$\sigma=50$}     & DIS    & \multirow{2}{*}{17.67}     & \multirow{2}{*}{24.12}     & 25.74     & 26.39 & 26.41 \\ \cline{2-2} \cline{5-7}
			& IIS    &                        &                        & 25.82    & 26.68     & 26.71     \\ \hline
	\end{tabular}}
\end{table}

We experiment the composited bilateral filter (C-BF) with FB of magnitude $3$, $9$ and $16$ generated by DIS and IIS respectively. These models are tested in BSD68 with Gaussian noise of standard variance $\sigma=25$ and $50$. As shown in Table \ref{tab:tab1}, small-scale FB generated by IIS performs better in both noise levels. When the magnitude of FB increases, the performance of C-BF tends to grow slowly, but shows better robustness under different inputs. To achieve a proper balance between performance, robustness and efficiency, we suggest to apply FB of magnitude $9$ produced by IIS for C-BF denoiser in real scenarios.

\textbf{Remark.} Conventional filters can only generate one optimal result using a single group of parameters. In contrast, our method learns the optimal composition of a series of filtering results that covers the distinctive effects of the original filter. That is the reason why the final result of our method outperforms the best filtering result in FB and also why fully connected layers for composition learning can be effective in this problem.
The construction of FB requires domain knowledge and tuning trials to decide the proper effective range. This step is originally implemented by user's repetitive trials, but now integrated in an offline procedure. At the testing stage, the computation of FB leads to more operating time than directly using single filters. As analyzed in Section \ref{sec:4.3}, the testing efficiency is mainly decided by the original image operator, including the operating time of single filters and the number of filters for constructing FB. As shown in Table \ref{tab:tab1}, the performance curve is flattening when the number of filters becomes large (PSNR: -0.02dB/ +0.03dB from number 9 to 16, $\sigma=25$ and $50$). Therefore, a reasonable approximation can be made with a small-scale FB, which may be preferred on resource constrained platforms. Also, compared with manual parameter tuning, the computations of different elements in FB are totally independent, therefore being easily parallelized.

\begin{figure*}[!th]
	\centering{
		\subfigure[Noisy Images]{
			\begin{minipage}[b]{0.18\textwidth}
				\includegraphics[width=\textwidth]{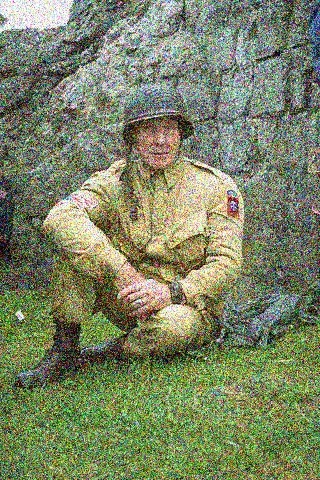} \\ 	
				\includegraphics[width=\textwidth]{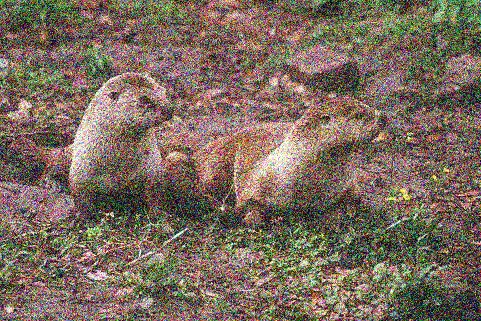}
		\end{minipage}}		
		\subfigure[Best in BFB]{
			\begin{minipage}[b]{0.18\textwidth}
				\includegraphics[width=\textwidth]{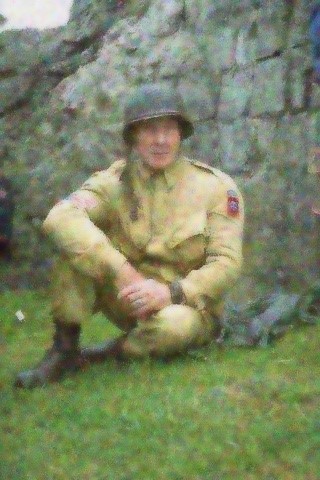} \\	
				\includegraphics[width=\textwidth]{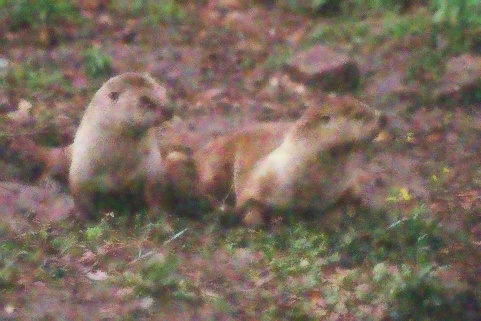}
		\end{minipage}}	
		\subfigure[Best in 1096]{
			\begin{minipage}[b]{0.18\textwidth}
				\includegraphics[width=\textwidth]{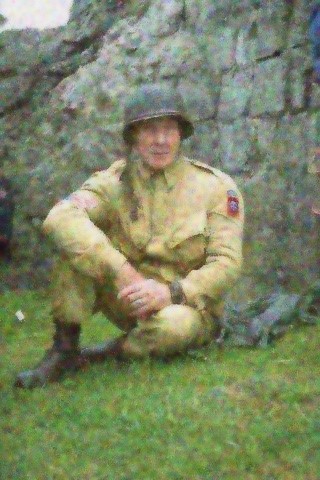} \\
				\includegraphics[width=\textwidth]{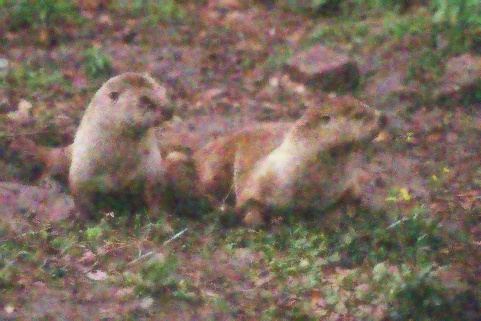}
		\end{minipage}}	
		\subfigure[C-BF]{
			\begin{minipage}[b]{0.18\textwidth}
				\includegraphics[width=\textwidth]{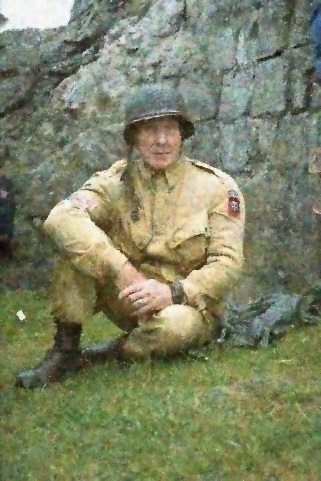} \\			
				\includegraphics[width=\textwidth]{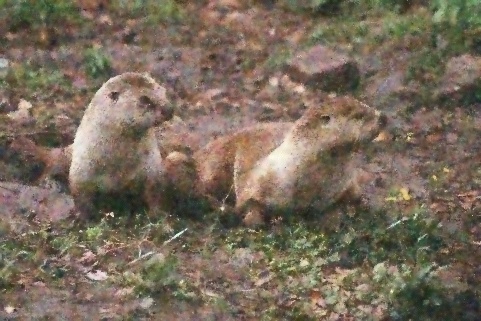}
		\end{minipage}}		
		\subfigure[Clean Images]{
			\begin{minipage}[b]{0.18\textwidth}
				\includegraphics[width=\textwidth]{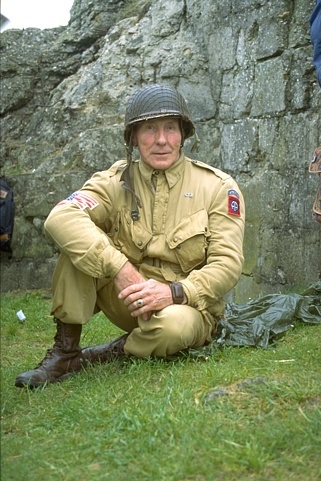} \\			
				\includegraphics[width=\textwidth]{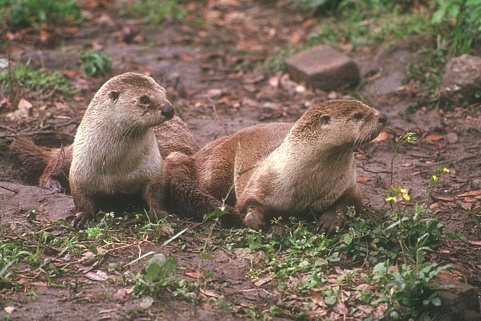}
		\end{minipage}}	
		
		\caption{Visual comparisons of image denoising. (a) Input; (b)-(c) Results of the highest PSNR in bilateral filtered basis and 1096 candidates; (d) Ours; (e) Noise-free images.}
		\label{fig:fig4} 
	}
\end{figure*}
\begin{figure*}[!th]
	\centering{
		\subfigure[Noisy Images]{
			\begin{minipage}[b]{0.18\textwidth}
				\includegraphics[width=\textwidth]{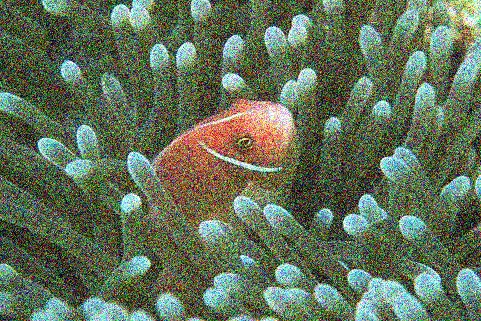} \\ 
				\includegraphics[width=\textwidth]{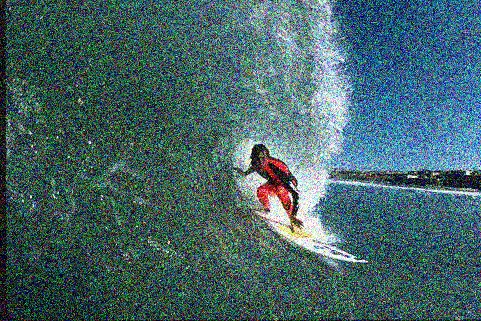} \\ 
				\includegraphics[width=\textwidth]{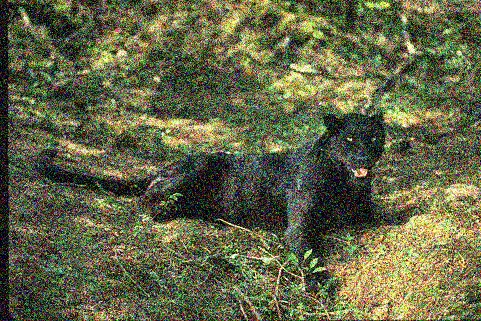} \\ 	
				\includegraphics[width=\textwidth]{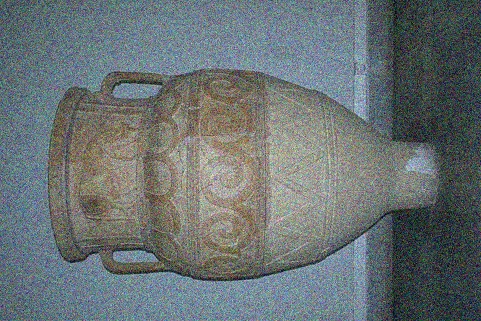}
		\end{minipage}}		
		\subfigure[Best in BFB]{
			\begin{minipage}[b]{0.18\textwidth}
				\includegraphics[width=\textwidth]{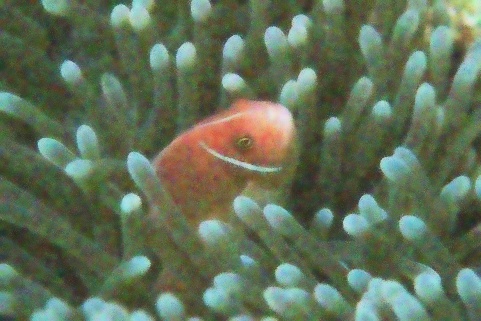} \\
				\includegraphics[width=\textwidth]{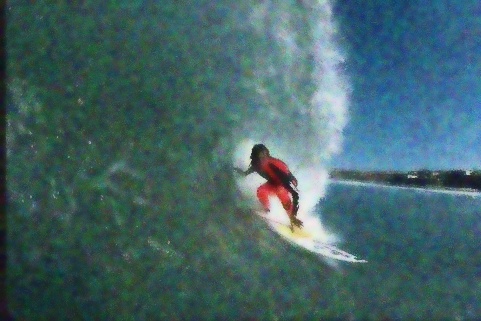} \\
				\includegraphics[width=\textwidth]{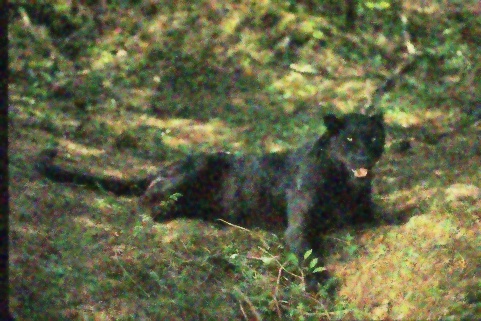} \\	
				\includegraphics[width=\textwidth]{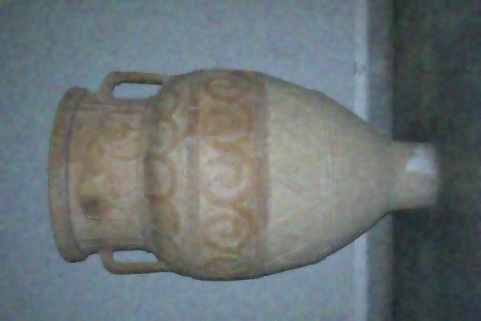}
		\end{minipage}}	
		\subfigure[Best in 1096]{
			\begin{minipage}[b]{0.18\textwidth}
				\includegraphics[width=\textwidth]{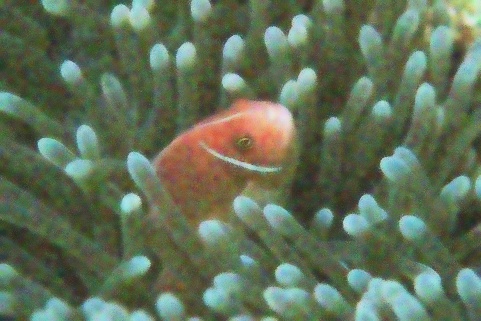} \\
				\includegraphics[width=\textwidth]{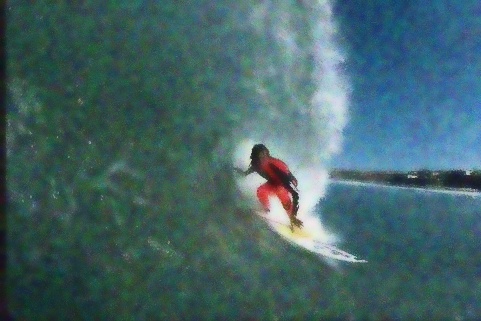} \\
				\includegraphics[width=\textwidth]{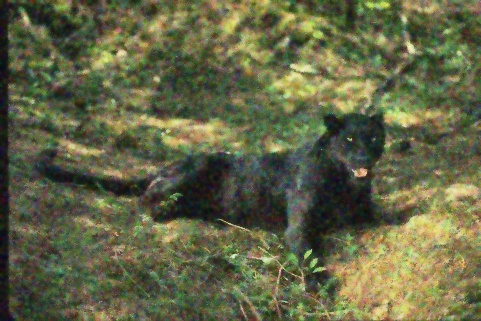} \\
				\includegraphics[width=\textwidth]{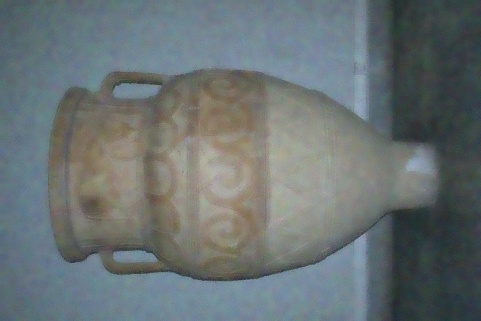}
		\end{minipage}}	
		\subfigure[C-BF]{
			\begin{minipage}[b]{0.18\textwidth}
				\includegraphics[width=\textwidth]{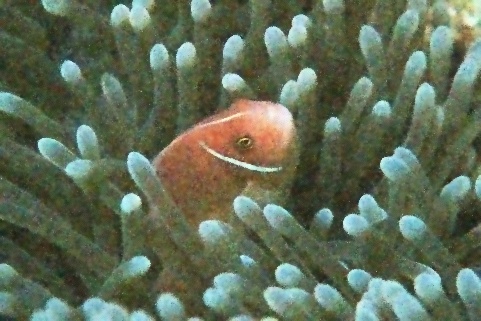} \\	
				\includegraphics[width=\textwidth]{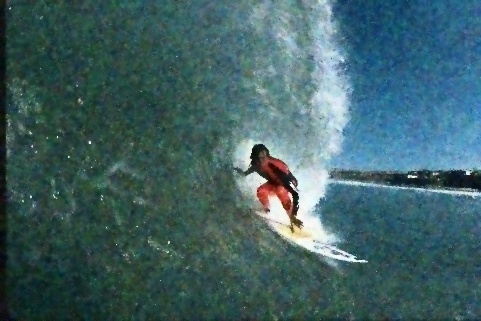} \\	
				\includegraphics[width=\textwidth]{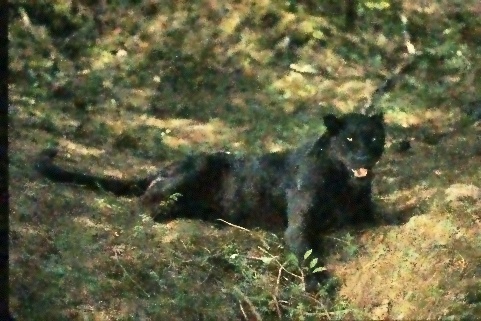} \\			
				\includegraphics[width=\textwidth]{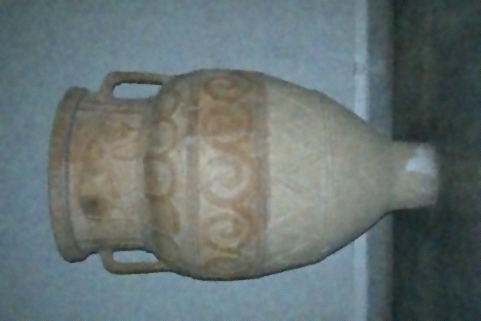}
		\end{minipage}}		
		\subfigure[Clean Images]{
			\begin{minipage}[b]{0.18\textwidth}
				\includegraphics[width=\textwidth]{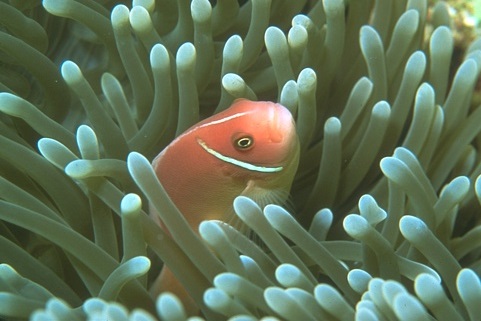} \\		
				\includegraphics[width=\textwidth]{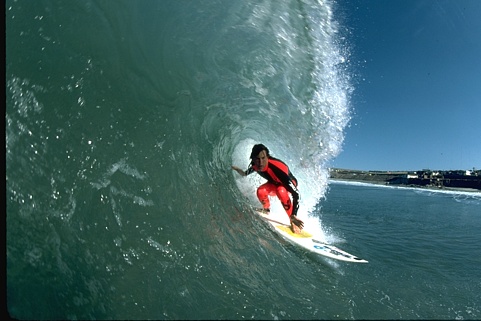} \\		
				\includegraphics[width=\textwidth]{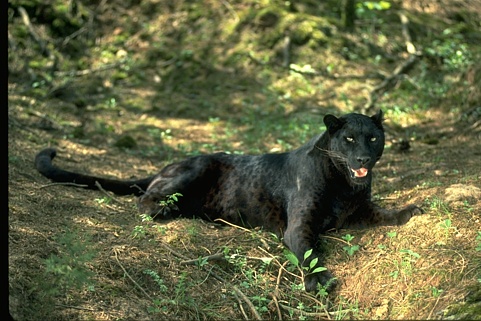} \\			
				\includegraphics[width=\textwidth]{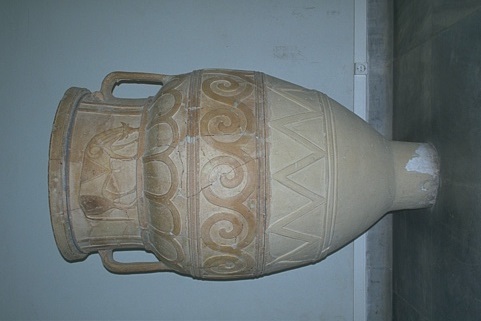}
		\end{minipage}}	
		
		\caption{More results of composited bilateral filter for image denoising.}
		\label{fig:bf_denoise} 
	}
\end{figure*}

\begin{figure}[!th]
	\centering{
		\subfigure[Input]{
			\begin{minipage}[b]{0.22\textwidth}
				\includegraphics[width=\textwidth]{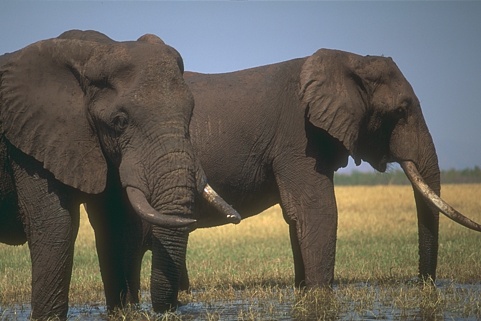} \\ 	
				\includegraphics[width=\textwidth]{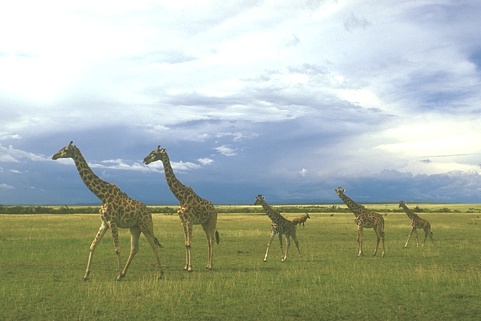}
		\end{minipage}}		
		\subfigure[C-BF]{
			\begin{minipage}[b]{0.22\textwidth}
				\includegraphics[width=\textwidth]{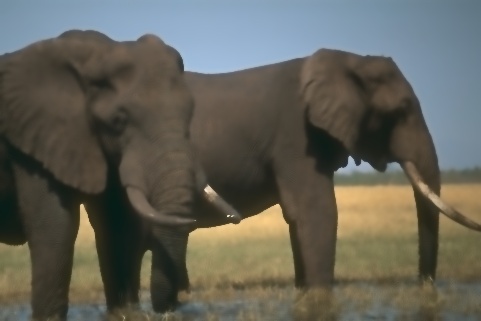} \\ 	
				\includegraphics[width=\textwidth]{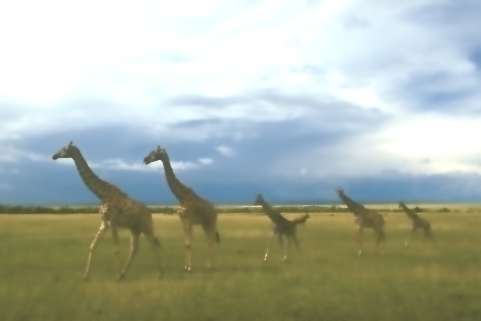}
		\end{minipage}}	
		\caption{ Generalization of C-BF on image smoothing.}
		\label{fig:bf_smooth} 
	}
\end{figure}

\section{Experimental Results} \label{sec4}

\subsection{Implementation on Popular Filters}

\textbf{Bilateral Filter}. We add different levels of Gaussian noise to BSD300 to train the composited bilateral filter (C-BF). MSE loss is applied in Eq. \ref{loss_t}. The bilateral filtered basis (BFB) is generated from 9 groups of parameters selected by Indirect Isometric Sampling (IIS).

Fig. \ref{fig:fig4} shows the comparison of C-BF with two reference images produced by the bilateral filter (BF). In case of subjectivity in manual parameter tuning, the reference images are chosen according to Peak Signal Noise Ratio (PSNR), which includes the best result in BFB and that tested in 1096 parameter sets by discretising the range of each parameter in a fine-grained way. It can be seen that C-BF outperforms the optimal approximation in BFB by a large margin, which shows the power of the proposed composite learning scheme. Moreover, after enumerating more than 1000 BF approximations, C-BF still yields better results. More results of composited bilateral filter for image denoising can be found in Fig. \ref{fig:bf_denoise}.

In addition, we also compare the results of our C-BF with BM3D \cite{bm3d}, TNRD \cite{tnrd} and DNCNN \cite{dncnn}. C-BF produces very comparable results with DNCNN \cite{dncnn} (PSNR: 30.22/26.68 vs. 30.98/26.98) on BSD68 with $\sigma=25$/$\sigma=50$, which are superior to BM3D \cite{bm3d} (PSNR: 29.66/26.32) and TNRD \cite{tnrd} (PSNR: 29.92/26.57)

\textbf{Remark.} Different from pure data-driven methods, in our framework, learning strategies are combined with analytic models from conventional wisdoms, which allows better generalization across different applications. An example is shown in Fig. \ref{fig:bf_smooth}: by directly applying the trained model to image smoothing, C-BF can smooth out image details (such as the texture on the lawn and the elephant's skin) and preserve salient edges without parameter tuning.
Therefore, C-BF trained on the denoising dataset also performs well on the image smoothing task.

\begin{figure*}[!th]
	\centering{
		\subfigure[Input]{
			\begin{minipage}[b]{0.22\textwidth}
				\includegraphics[width=\textwidth]{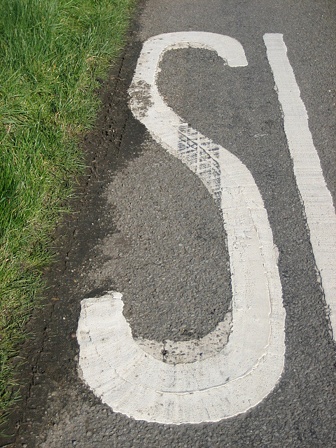} \\
				\includegraphics[width=\textwidth]{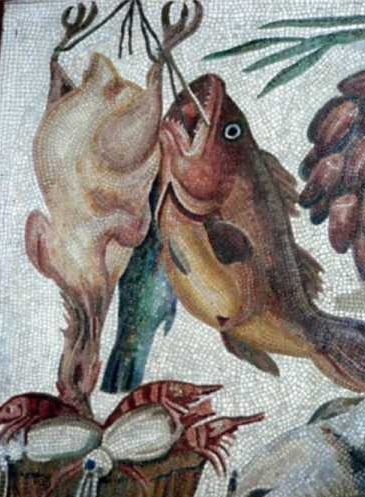}
		\end{minipage}}	
		\subfigure[RGF]{
			\begin{minipage}[b]{0.22\textwidth}
				\includegraphics[width=\textwidth]{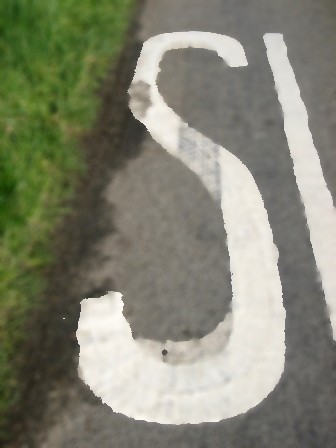} \\
				\includegraphics[width=\textwidth]{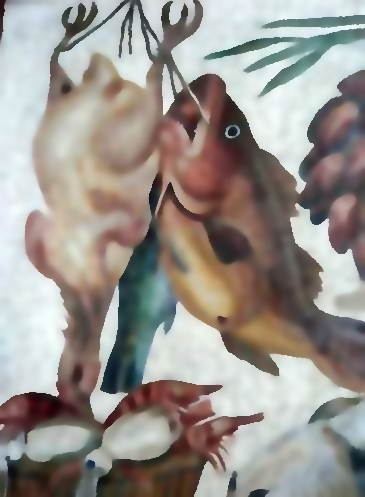}
		\end{minipage}}	
		\subfigure[C-RGF (ours)]{
			\begin{minipage}[b]{0.22\textwidth}
				\includegraphics[width=\textwidth]{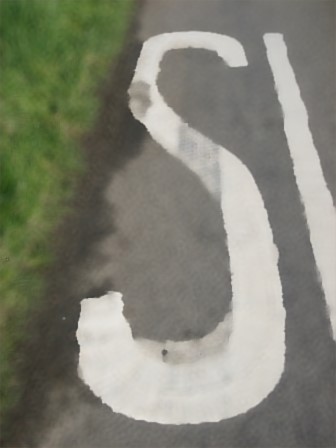} \\
				\includegraphics[width=\textwidth]{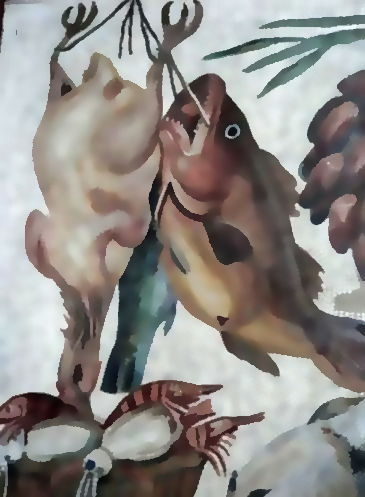}
		\end{minipage}}		
		\subfigure[RTV]{
			\begin{minipage}[b]{0.22\textwidth}
				\includegraphics[width=\textwidth]{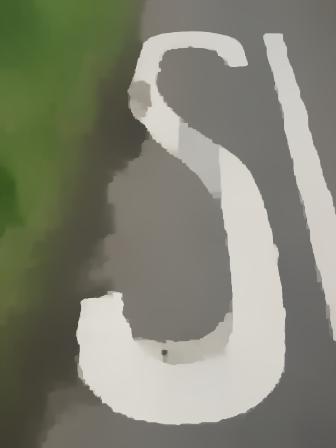} \\
				\includegraphics[width=\textwidth]{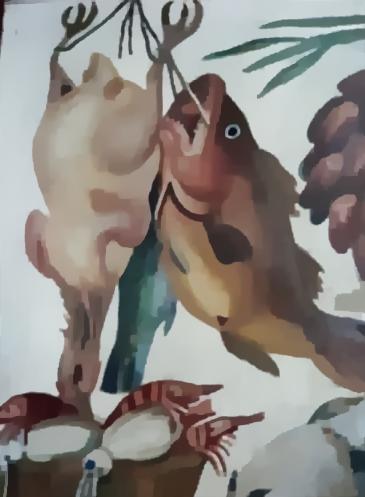}
		\end{minipage}}
		\caption{Visual comparisons of image texture removal.}
		\label{fig:fig2} 
	}
\end{figure*}

\begin{figure*}[!th]
	\centering{
		\subfigure[Input]{
			\begin{minipage}[b]{0.22\textwidth}
				\includegraphics[width=\textwidth]{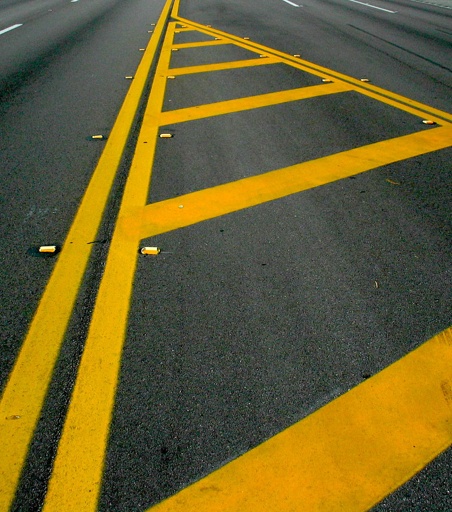} \\ 
				\includegraphics[width=\textwidth]{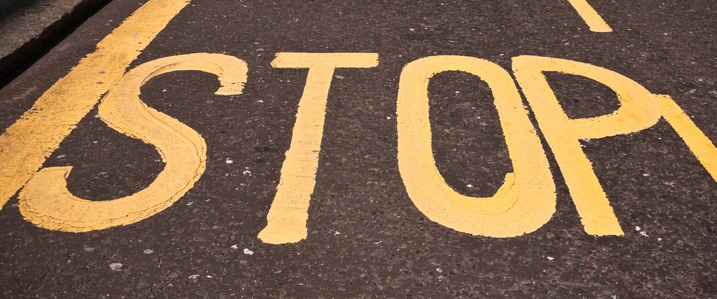} \\ 	
				\includegraphics[width=\textwidth]{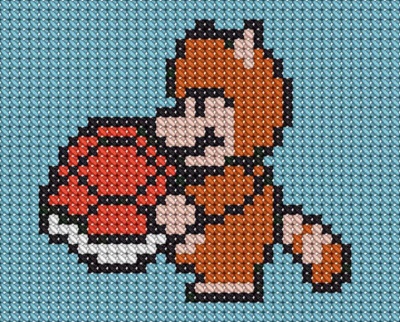}
		\end{minipage}}		
		\subfigure[RGF]{
			\begin{minipage}[b]{0.22\textwidth}
				\includegraphics[width=\textwidth]{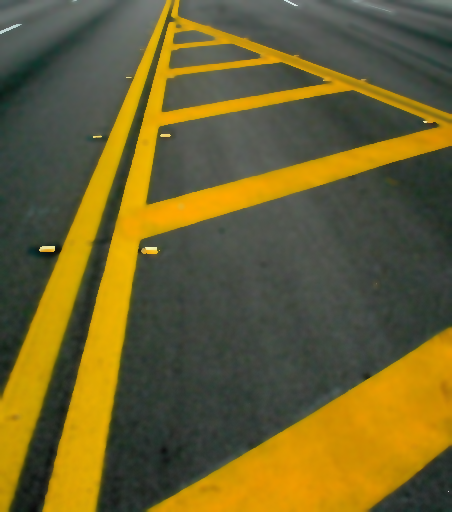} \\
				\includegraphics[width=\textwidth]{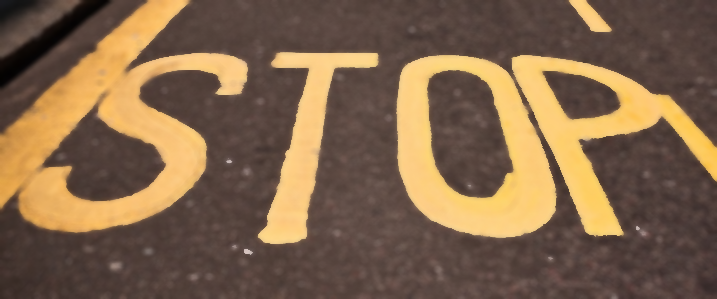} \\	
				\includegraphics[width=\textwidth]{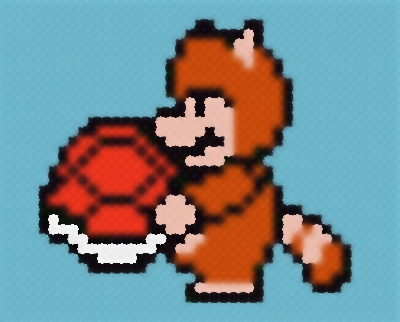}
		\end{minipage}}	
		\subfigure[C-RGF (ours)]{
			\begin{minipage}[b]{0.22\textwidth}
				\includegraphics[width=\textwidth]{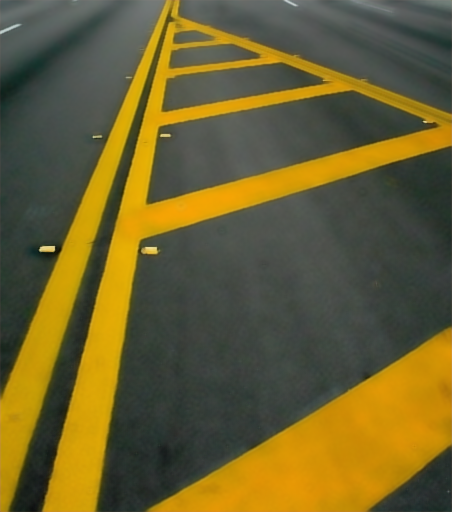} \\
				\includegraphics[width=\textwidth]{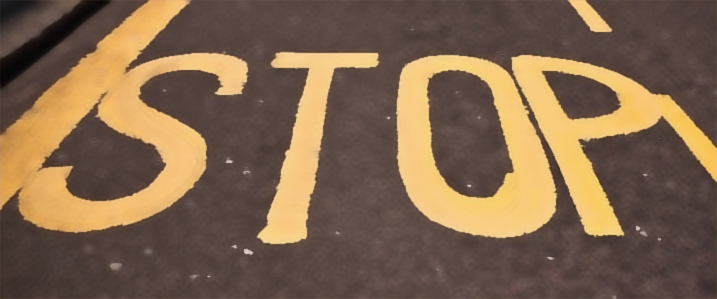} \\
				\includegraphics[width=\textwidth]{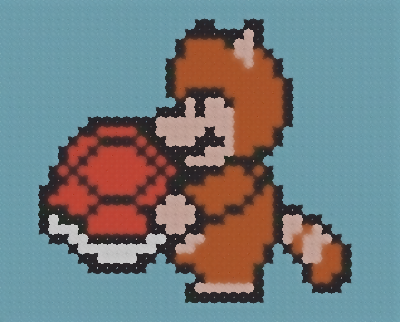}
		\end{minipage}}	
		\subfigure[RTV]{
			\begin{minipage}[b]{0.22\textwidth}
				\includegraphics[width=\textwidth]{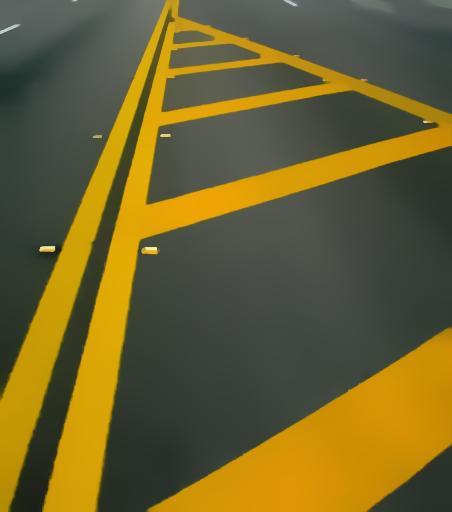} \\	
				\includegraphics[width=\textwidth]{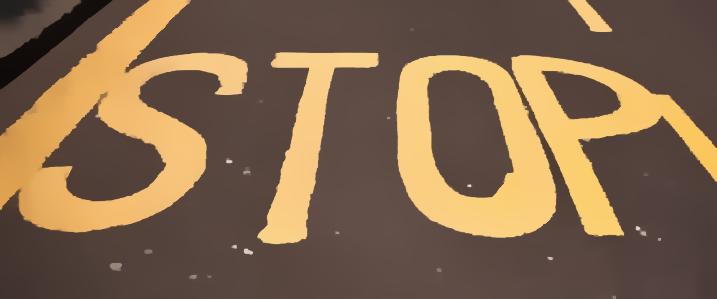} \\			
				\includegraphics[width=\textwidth]{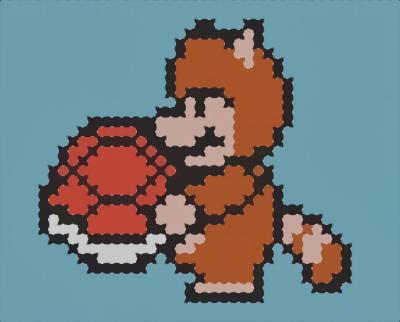}
		\end{minipage}}		
		
		\caption{More results of image texture removal.}
		\label{fig:rgf_detexture} 
	}
\end{figure*}

\textbf{Rolling Guidance Filter.}
Rolling Guidance Filter (RGF) \cite{RGF} is a popular choice for image smoothing. RGF applies the Gaussian Filter of scale $\sigma_{s}$ to generate a guidance image with eliminated details and then uses bilateral filter to iteratively recover image structures. Denote the iteration number as $t$, 8 groups of parameters are chosen by isometric parameter sampling to construct the Rolling Guidance Filtered Basis (RGFB), where $\sigma_{c}=\{0.2,0.5\}$, $\sigma_{s}=\{3,6\}$, $k = 9$, $t = \{2,4\}$.

Concerning the image smoothing problem, superior visual effects can be achieved by time-consuming optimization based approaches. Accordingly, RTV \cite{RTV}, which is a representative optimization based method with a well-defined texture metric, is applied to build ground-truth images for  RGF. In Eq. \ref{loss_t}, $L_{1}$ loss is applied and an additional Total Variation loss is used to constrain the smoothness of the final result.

To compare C-RGF with manually tuned RGF, we provide 5 candidates of RGF results for 20 participants to rate and then exhibit the one with highest average score. As shown in Fig. \ref{fig:fig2}, C-RGF finds the best potential of RGF to balance between structure preserving and texture removal. More results of image smoothing can be found in Fig. \ref{fig:rgf_detexture}.

\begin{figure*}[!th] 
	\centering{
		\subfigure[Noisy Images]{
			\begin{minipage}[b]{0.23\textwidth}
				\includegraphics[width=\textwidth]{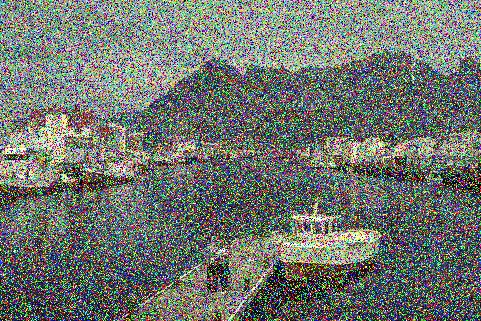} \\ 
				\includegraphics[width=\textwidth]{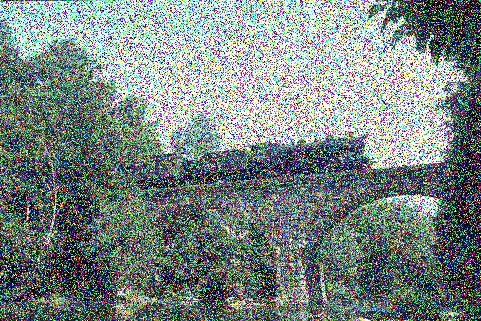}\\
				\includegraphics[width=\textwidth]{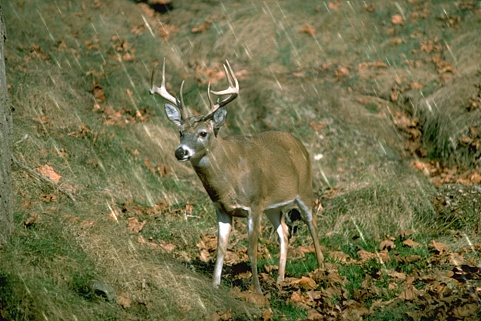}
		\end{minipage}}
		\subfigure[MF]{
			\begin{minipage}[b]{0.23\textwidth}
				\includegraphics[width=\textwidth]{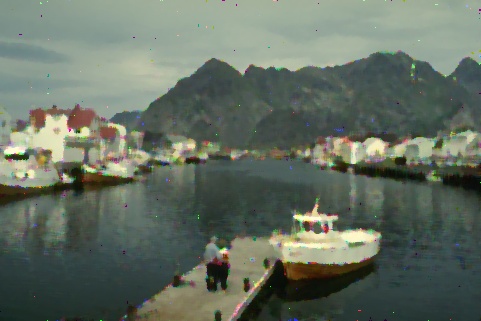} \\
				\includegraphics[width=\textwidth]{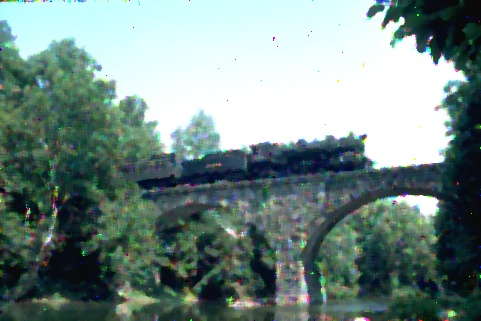}\\
				\includegraphics[width=\textwidth]{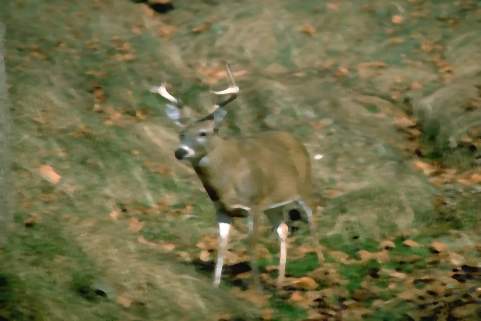}
		\end{minipage}}
		\subfigure[C-MF (ours)]{
			\begin{minipage}[b]{0.23\textwidth}
				\includegraphics[width=\textwidth]{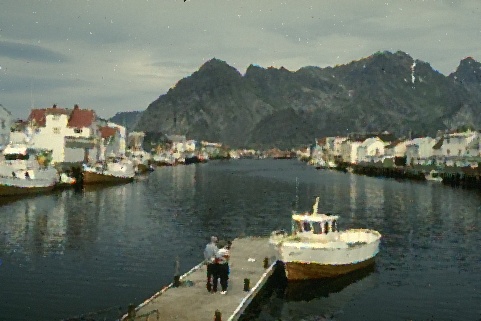} \\ 
				\includegraphics[width=\textwidth]{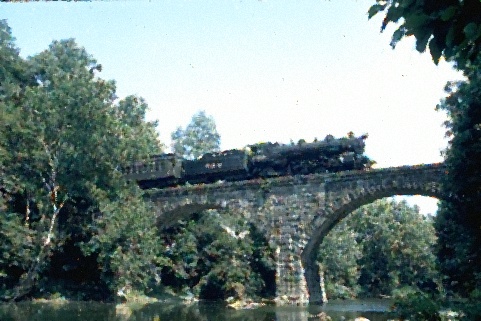}\\
				\includegraphics[width=\textwidth]{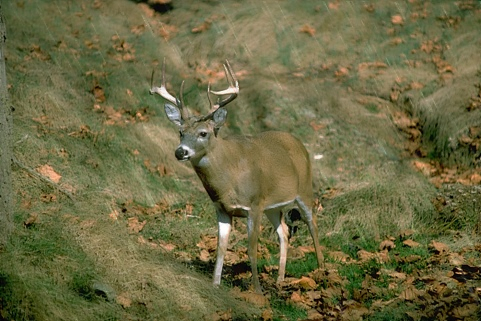}
		\end{minipage}}
		\subfigure[Clean Images]{
			\begin{minipage}[b]{0.23\textwidth}
				\includegraphics[width=\textwidth]{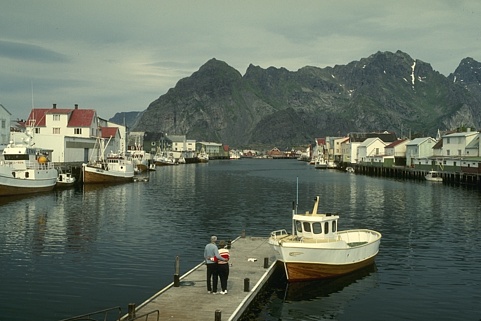} \\ 
				\includegraphics[width=\textwidth]{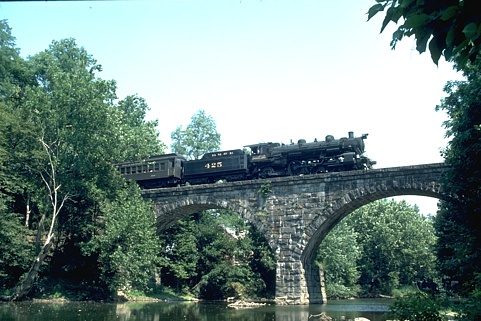}\\
				\includegraphics[width=\textwidth]{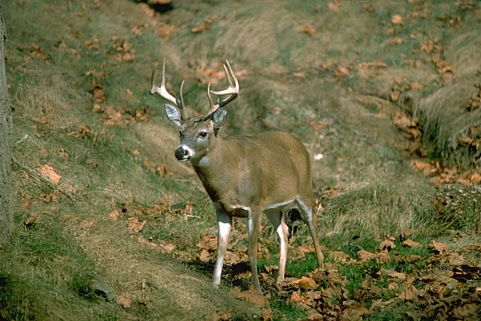}
		\end{minipage}}
		\caption{Visual comparisons of impulse noise removal and image deraining. 
		}
		\label{fig:med}}
\end{figure*}

 \begin{figure*}[!th] 
	 	\centering{
		 		\subfigure[Noisy Images (Input)]{
			 			\begin{minipage}[b]{0.3\textwidth}
				 				\includegraphics[width=\textwidth]{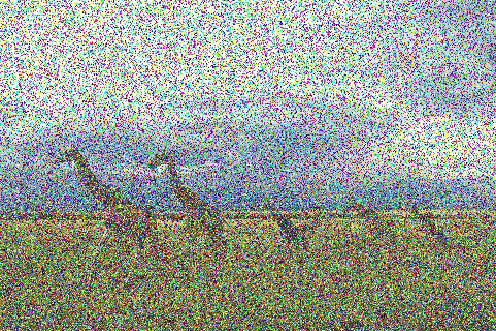} \\
				 				\includegraphics[width=\textwidth]{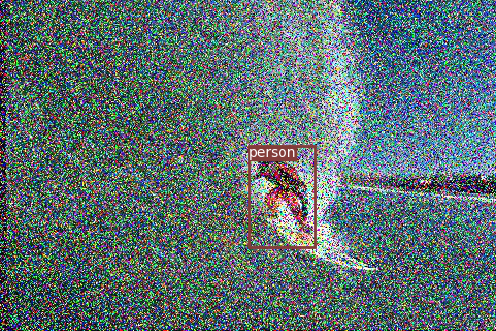}\\
				 				\includegraphics[width=\textwidth]{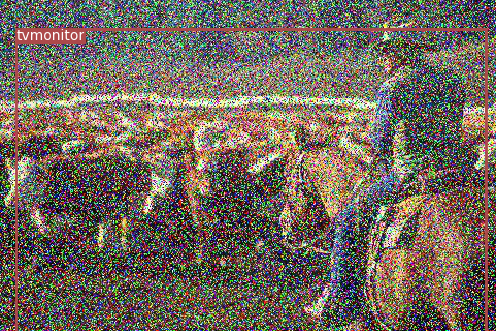} \\ 
				 				\includegraphics[width=\textwidth]{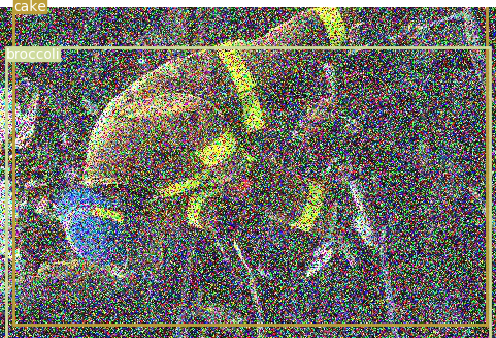} \\ 
				 				\includegraphics[width=\textwidth]{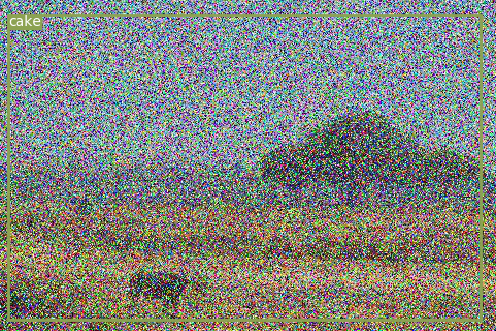}
							\end{minipage}}
					\subfigure[after MF]{
			 				\begin{minipage}[b]{0.3\textwidth}
				 					\includegraphics[width=\textwidth]{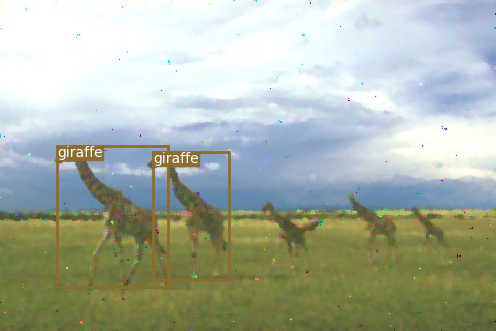} \\
									\includegraphics[width=\textwidth]{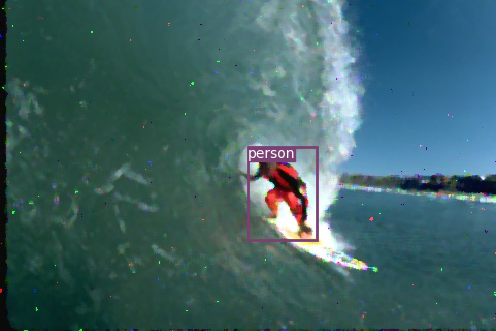}\\
				 					\includegraphics[width=\textwidth]{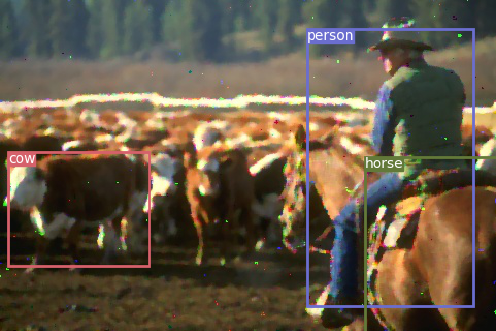} \\
				 					\includegraphics[width=\textwidth]{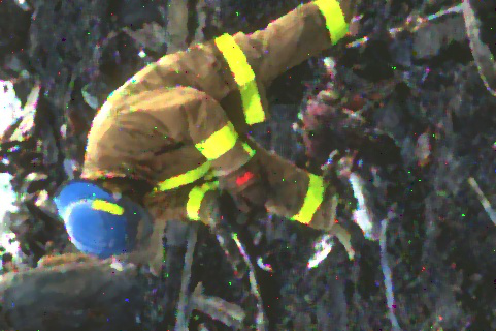} \\	
				 					\includegraphics[width=\textwidth]{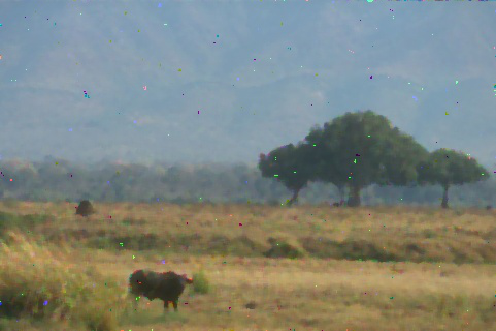}
				 				\end{minipage}}
		 				\subfigure[after C-MF]{
			 					\begin{minipage}[b]{0.3\textwidth}
				 						\includegraphics[width=\textwidth]{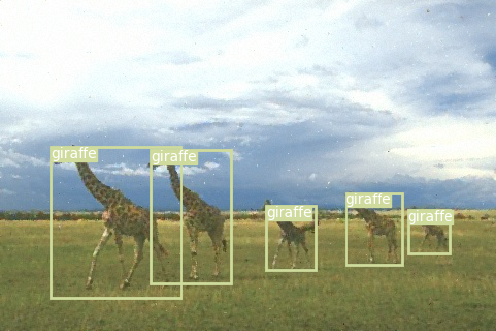} \\
				 						\includegraphics[width=\textwidth]{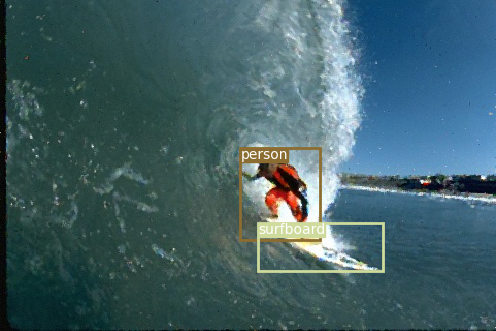}\\
				 						\includegraphics[width=\textwidth]{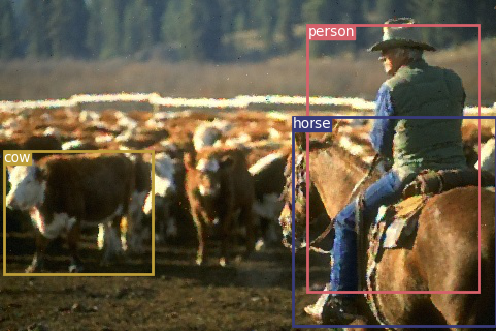} \\
				 						\includegraphics[width=\textwidth]{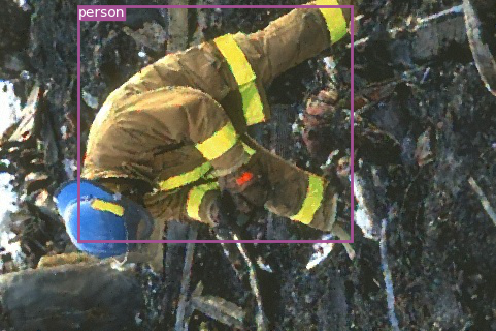} \\
				 						\includegraphics[width=\textwidth]{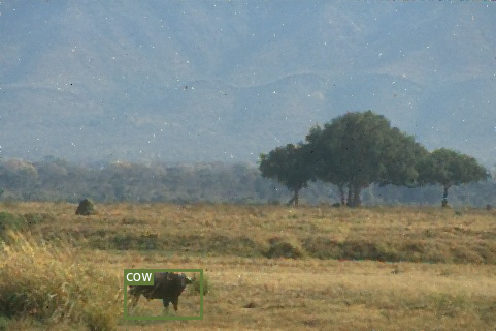}
				 					\end{minipage}}
		 					\caption{Object detection results after performing median filter and its parafree version.}
		 					\label{fig:yolo}}
				\end{figure*}

 \begin{figure*}[!th] 
	 	\centering{
				\subfigure[Original Images]{
						\begin{minipage}[b]{0.3\textwidth}
								\includegraphics[width=\textwidth]{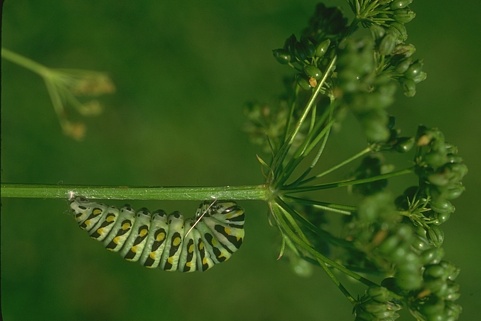} \\
				 				\includegraphics[width=\textwidth]{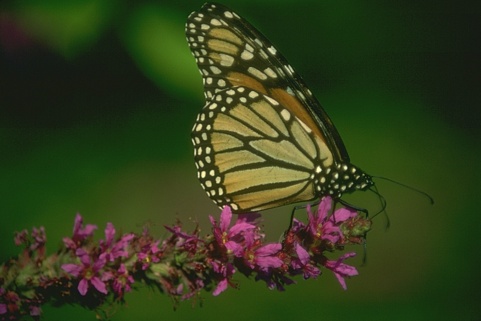}
				 			\end{minipage}}
					\subfigure[After RGF]{
			 				\begin{minipage}[b]{0.3\textwidth}
				 					\includegraphics[width=\textwidth]{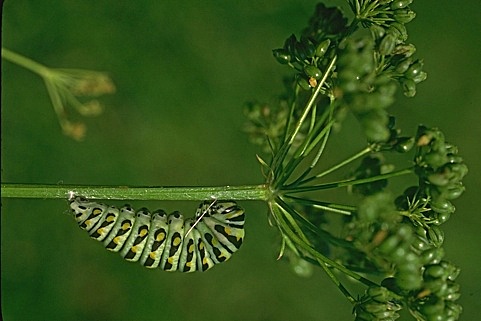} \\
				 					\includegraphics[width=\textwidth]{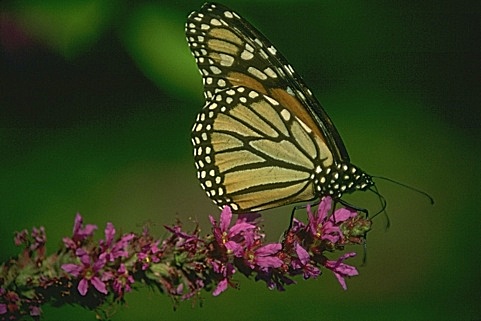}
				 				\end{minipage}}
		 				\subfigure[After C-RGF]{
			 					\begin{minipage}[b]{0.3\textwidth}
				 						\includegraphics[width=\textwidth]{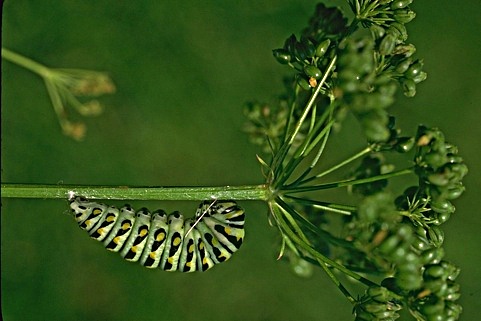} \\
				 						\includegraphics[width=\textwidth]{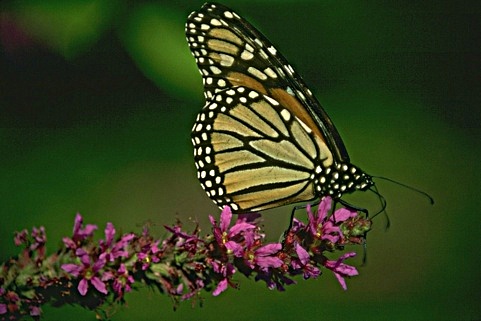}
				 					\end{minipage}}
		 					\caption{Detail enhancement results after performing rolling guidance filter and its basis compositin learning version.}
		 					\label{fig:detail}}
	 			\end{figure*}

\textbf{Median Filter.}
Median filter (MF) is a powerful tool to recover extremely biased pixels. Different levels of salt-and-pepper noise is added to BSD300 to train the composited median filter (C-MF). MSE loss is applied in Eq. \ref{loss_t}. We use MF of 8 different window shapes $k_{1}*k_{2}$ to generate median filtered basis (MFB), i.e., $(k_{1},k_{2})=(3,3),(3,5),(3,7),(3,9),(5,5),(5,7),\\ (5,9),(7,7)$. As shown in the first row of Fig. \ref{fig:med}, C-MF recovers the image contaminated by $40\%$ impulse noise smoothly, without the artifacts produced by the original MF.

Furthermore, we apply our model on single image deraining. Single image deraining is a challenging problem compared to video-based rain removal due to the lack of inter-frame information. Thus, the mainstream in this area depends on sophisticated designed prior or deep neural networks. However, we demonstrate that a simple MF under the composition learning framework can also benefit the challenging deraining task.

An important prior knowledge of rain streaks is that they look brighter than background pixels \cite{garg2007vision}. As suggested in the last row of Fig. \ref{fig:med}, MF is able to suppress rain streak pixels but suffers from losing details. This problem is substantially alleviated by its composited version.

\textbf{Remark.} We propose basis composition learning to optimize the performance and user experience of parameterized image operators. In Section \ref{sec4}, we only give examples on classical and simple image operators to focus more on the effectiveness of the proposed framework rather than the operator itself. Better performance can be achieved by using more complex or specialized filters, but might cost more in the testing time. An example is C-CBM3D for denoising (PSNR: 31.40dB/28.07dB, $\sigma=25$/$\sigma=50$). Moreover, the filtered basis (FB) can serve as a set of effective features that benefits not only our method but also other learning based approaches. As discussed above, FB introduces domain knowledge on specific tasks and thus eases the subsequent learning process. We experiment DNCNN facilitated with bilateral filtered basis and an performance increase can be observed in Table \ref{tab:tab3}.

 \begin{figure}[]
	 	\centering{
		 		\subfigure[Noisy Image]{			
			 			\includegraphics[width=0.44\linewidth]{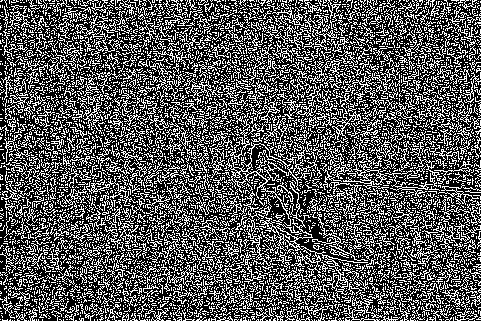}}
		 		\subfigure[After BF]{
			 			\includegraphics[width=0.44\linewidth]{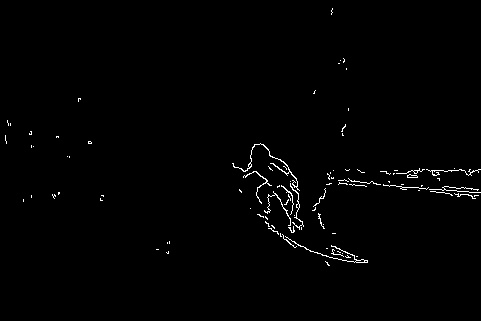}}
		 		\subfigure[After C-BF]{
			 			\includegraphics[width=0.44\linewidth]{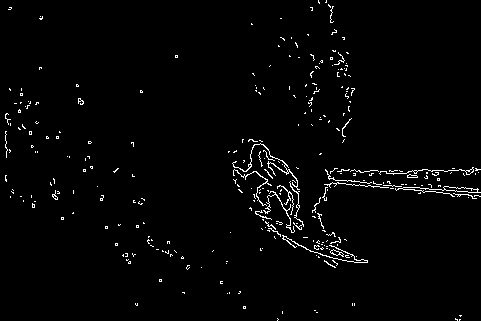}}
		 		\subfigure[Original Image]{
			 			\includegraphics[width=0.44\linewidth]{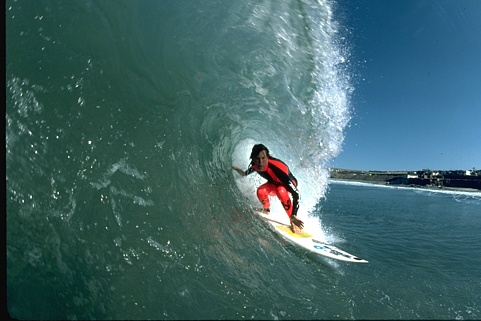}}
		 		\caption{Edge extraction results after performing bilateral filter and its basis composition learning version.}
		 		\label{fig:canny} 
			}
	\end{figure}

 \subsection{Applications}
 Filtering is an indispensable pre-processing step in many computer vision applications. In this section, we demonstrate the results of object detection, edge extraction and detail enhancement after conducting the proposed basis composition learning solution, which outperforms the original filters without parameter tuning.

\textbf{Object Detection.}
 The performance of object detection can be greatly degraded when the images are corrupted by impulsive noise. We adopt the trained model of YOLOv3 \cite{redmon2018yolov3} (provided by the authors) on the images contaminated by $40\%$ impulse noise and find that the objects can hardly be recognized (see Fig. \ref{fig:yolo} (a)). Then, we detect on images filtered by median filter (MF) and basis composition learning version (C-MF), respectively. As shown in Fig. \ref{fig:yolo} (b)\&(c), C-MF smoothly removes the noise without generating artifacts or losing the image details, which allows more accurate detection results. 

 \textbf{Edge Extraction.}
 The edges of an image are semantically important but cannot be extracted when interfered by Gaussian noise (see Fig. \ref{fig:canny} (a)). This problem can be alleviated after using bilateral filter. As shown in Fig. \ref{fig:canny}, we use Canny operator to detect edges on the results of the original bilateral filter (BF) and the C-BF. After denoised by C-BF, the extracted edge map is more semantically close to the original image due to better preservation of image structures.

 \textbf{Detail Enhancement.}
 Image texture removal can benefit detail enhancement of natural images by inversely adding back the texture residuals. As suggested in Fig. \ref{fig:detail}, since the basis composition learning of rolling guidance filter (C-RGF) automatically finds the best potential of rolling guidance filter (RGF), it better maintains the main image structures and flattens the image textures, which leads to more accurate enhancement of image details and more visual appealing effects.

\subsection{Computational Efficiency} \label{sec:4.3}
The computational cost of composited filters mainly depends on the construction of FB. When using FB of magnitude $K$, the complexity of composited filters is $O(K*f+c)$, where $f$ represents the complexity of BF and $c$ is the cost of composition module (nearly negligible). With smaller FB, the approximation of composited filters can be made to trade accuracy for efficiency, which may be preferred on resource constrained platforms. Take C-BF for instance, when the parameter sets ($\sigma_{c},\sigma_{s},k$) are set as (0.1, 2, 15), (0.5, 1.5, 15) and (1, 3, 15), the performance is only moderately reduced to an average PSNR of 29.98. A reasonable approximation can be made with a small-scale FB. For further acceleration, there exist many algorithms to speed up the classical filters, such as permutohedral lattice technique \cite{adams2010fast} for bilateral filter. Also, the computations of different elements in FB are totally independent, therefore they can be easily parallelized.

\subsection{Comparison with The State-of-The-Arts}
Note that \textit{our goal is to provide a general composition learning framework for image filters, rather than being the top in specific applications.} 
The basic network can be deepened to boost performance at the cost of memory efficiency. Even so, we provide comparisons with many previous works to demonstrate that our general scheme generates comparative results in single tasks using simple filters.

For image denoising, C-BF is compared with BM3D \cite{bm3d}, TNRD \cite{tnrd} and DNCNN \cite{dncnn}. C-BF produces very comparable results with DNCNN \cite{dncnn} (PSNR: 30.22/26.68 vs. 30.98/26.98) on BSD68 with $\sigma=25$/$\sigma=50$, which are superior to BM3D \cite{bm3d} (PSNR: 29.66/26.32) and TNRD \cite{tnrd} (PSNR: 29.92/26.57).
For image deraining, we compare C-MF with popular conventional methods: UGSM \cite{deng2018directional}, GMM \cite{li2016rain}, and state-of-the-art deep-learning based methods: RESCAN \cite{li2018recurrent}, DRD \cite{deng2019drd}. Tested on Rain100L, C-MF (PSNR/SSIM: 28.54/0.93) outperforms both GMM (PSNR/SSIM: 27.16/0.89) and UGSM (PSNR/SSIM: 27.32/0.0.90). DRD (PSNR/SSIM: 37.15/0.98) and RESCAN (PSNR/SSIM: 37.07/0.98) have advantages in these metrics using much heavier networks than C-MF.

\section{More Discussions} \label{sec5}
\subsection{The Effect of Residual Branch}
Problems in applications of image filters generally follow the additive composition model: $I = J + R$, where $R$ denotes the residual between the original image $I$ and the filtering result $J$. For instance, $R$ refers to noise in image denoising and details in image smoothing. This enables the design of a dual-branch composition module where the residual branch learns from the FB's residual to the ground-truth $R$. In this way, the two branches learn the same target but with different data, which benefits the final performance with little cost.

To validate the effectiveness of the residual branch, we compared the qualitative results of composited bilateral filter (C-BF) only with the content branch and that with the dual branch composition module, based on different magnitudes of filtered basis (FB). Table \ref{tab:tab2} shows that C-BF with a residual branch generates better results and this improvement is more obvious when the magnitude of FB reduces. This is mainly because that the residual branch helps to find back the lost structure components in FB, which not only improves the final performance, but also benefits a better trade-off between the efficiency and effectiveness.

\begin{table}[h]
	\small
	\centering{
		\caption{\small Ablation study on the residual branch (PSNR).}
		\label{tab:tab2}
		\vspace{5pt}
		\begin{tabular}{|c|c|c|c|}
			\hline
			\multicolumn{2}{|c|}{Magnitude of FB} & 3     & 9     \\ \hline
			\multirow{2}{*}{$\sigma=25$}   & w/      & 29.98 & 30.14 \\ \cline{2-4}
			& w/o   & 29.24 & 29.92 \\ \hline
			\multirow{2}{*}{$\sigma=50$}   & w/      & 25.82     & 26.68 \\ \cline{2-4}
			& w/o   & 25.16     & 26.35     \\ \hline
	\end{tabular}}
\end{table}

\begin{table}[h]
	\footnotesize
	\centering{
		\caption{Comparisons of denoising performance and network magnitudes. \small }
		\label{tab:tab3}
		\vspace{5pt}
		\begin{tabular}{|c|c|c|c|}
			\hline
			Method                       & C-BF & C-DNCNN & DNCNN \\ \hline
			PSNR                         & 29.16   & 30.08  & 29.72     \\ \hline
			Network Parameters 			 & 189     & 559K     & 559K     \\ \hline
	\end{tabular}}
\end{table}

 	\subsection{The Magnitude of Network} \label{sec5.2}
 Through extensive experiments, we have demonstrated that the effort of basis composition learning for image filters produces better approximations than the original versions. This is realized when the basic network in the composition module is limited to a very shallow fully connected network. In this circumstance, our goal is to outperform the manually tuned filter without tedious parameter tuning but still remain its simplicity. The success attributes to the design of FB, which integrates the domain-specific knowledge to build effective input features highly correlated to the target. It is apparent that a more sophisticated network brings better results. To further demonstrate the effectiveness of the proposed FB, we have experimented to deepen the basic network in the composition module to be as the same magnitude of existing deep networks. One example is shown in Table \ref{tab:tab3}, we combine the bilateral filtered basis (BFB) with the deep network in DNCNN \cite{dncnn} (denoted as C-DNCNN). Tested in BSD68 with $\sigma = 30$, C-DNCNN outperforms the original DNCNN when sharing the same hyper-parameters and training strategy.

\subsection{Difference to Filter Learning}
Recently, several methods \cite{xu2015deep,fan2017generic,chen2017fast} realized a similar purpose of `parameter-free' filtering. These methods aim to approximate and accelerate conventional image operators in a general framework. The filtering effects are learned from training pairs where the filtering result with fixed parameters is regarded as the ground truth. However, the problem is that filters with fixed parameters do not always suit the varying inputs, which means that no matter how deep the network is, it only approximates the non-optimal filtering effect. In contrast, the proposed method solve this problem by considering the filter itself rather than building datasets of filtering effects. The generality of our method comes from the construction of FB, which accommodates a range of possibilities in applying filters. Moreover, filter learning approaches claim to speed up image operators but using deep networks with massive parameters, which is storage-consuming in practice. In contrast, composited filters benefit from a lightweight learning scheme with much less storage burden.

\subsection{Generalization}
Essentially, basis composition learning is a general framework for any parameterized image operators. As long as a group of parameters is selected, the filtered basis (FB) can be generated for successive approximation. However, for time-consuming operators, such as some global optimization-based methods, the computational complexity of their composited version is propotional to the magnitude of FB, which is prohibitive in practice. This limitation is the main reason why this paper narrows the range of study to efficient image filters.

\section{Conclusion} \label{sec6}
In this paper, we present a generic basis composition learning paradigm to enhance the usability and performance of image filters. Prior knowledge and two different sampling methods are combined to select robust approximations for building the filtered basis. A dual-branch composition learning module accepts the filtered basis and their residuals as input features, where fully connected layers are leveraged to learn the basis composition coefficients along the channels. Experimental results on several applications validate the effectiveness of our method. In future, we will attempt to extend this approach to 3D surfaces. 


{\small
    \bibliographystyle{spbasic}
    \bibliography{adafil}
}

\end{document}